\documentclass[conference]{IEEEtran}
\IEEEoverridecommandlockouts
\usepackage{gensymb}
\usepackage{multirow}
\usepackage[ruled,vlined,linesnumbered]{algorithm2e}
\usepackage{url}

\usepackage{cite}
\usepackage{amsmath,amssymb,amsfonts}
\usepackage{algorithmic}
\usepackage{graphicx}
\usepackage{textcomp}
\usepackage{xcolor}
\def\BibTeX{{\rm B\kern-.05em{\sc i\kern-.025em b}\kern-.08em
    T\kern-.1667em\lower.7ex\hbox{E}\kern-.125emX}}
\begin{document}

\title{Robust and Resource-efficient Machine Learning Aided Viewport Prediction in Virtual Reality\thanks{This publication was supported by the U.S. Naval Research Laboratory under Grant N00173-21-1-G006, U.S. National Science Foundation and The Academy of Finland under Award 2132573, and the Army Research Office under Agreement Number W911NF1810378.

Accepted for publication in 2022 IEEE International Conference on Big Data (IEEE BigData 2022).
}
}

\author{\IEEEauthorblockN{Yuang Jiang}
\IEEEauthorblockA{
\textit{Department of Electrical Engineering} \\
\textit{Yale University}\\
yuang.jiang@yale.edu}
\and
\IEEEauthorblockN{Konstantinos Poularakis}
\IEEEauthorblockA{
\textit{Department of Electrical Engineering} \\
\textit{Yale University}\\
konstantinos.poularakis@yale.edu}
\and
\IEEEauthorblockN{Diego Kiedanski}
\IEEEauthorblockA{
\textit{Department of Electrical Engineering} \\
\textit{Yale University}\\
diego.kiedanski@yale.edu}
\and
\IEEEauthorblockN{Sastry Kompella}
\IEEEauthorblockA{
\textit{Information Technology Division} \\
\textit{U.S. Naval Research Laboratory}\\
sastry.kompella@nrl.navy.mil}
\and
\IEEEauthorblockN{Leandros Tassiulas}
\IEEEauthorblockA{
\textit{Department of Electrical Engineering} \\
\textit{Yale University}\\
leandros.tassiulas@yale.edu}
}

\IEEEoverridecommandlockouts
% \IEEEpubid{\makebox[\columnwidth]{978-1-6654-8045-1/22/\$31.00~\copyright2022 IEEE \hfill} \hspace{\columnsep}\makebox[\columnwidth]{ }}
\maketitle
% \IEEEpubidadjcol

\begin{abstract}
360-degree panoramic videos have gained considerable attention in recent years due to the rapid development of head-mounted displays (HMDs) and panoramic cameras. One major problem in streaming panoramic videos is that panoramic videos are much larger in size compared to traditional ones. Moreover, the user devices are often in a wireless environment, with limited battery, computation power, and bandwidth. To reduce resource consumption, researchers have proposed ways to predict the users’ viewports so that only part of the entire video needs to be transmitted from the server. However, the \emph{robustness} of such prediction approaches has been overlooked in the literature: it is usually assumed that only a few models, pre-trained on past users’ experiences, are applied for prediction to all users. We observe that those pre-trained models can perform poorly for some users because they might have drastically different behaviors from the majority, and the pre-trained models cannot capture the features in unseen videos. In this work, we propose a novel meta learning based viewport prediction paradigm to alleviate the worst prediction performance and ensure the robustness of viewport prediction. This paradigm uses two machine learning models, where the first model predicts the viewing direction, and the second model predicts the minimum video prefetch size that can include the actual viewport. We first train two meta models so that they are sensitive to new training data, and then quickly adapt them to users while they are watching the videos. Evaluation results reveal that the meta models can adapt quickly to each user, and can significantly increase the prediction accuracy, especially for the worst-performing predictions.
\end{abstract}


\section{Introduction}
360-degree videos, also known as panoramic  or immersive videos, are videos where a view in every direction is recorded at the same time. During playback, a video watcher has control of the viewing direction in real-time within the video panorama, and the visible area determined by the viewing direction is known as the \textit{viewport}.

As an important application of virtual reality (VR), 360\degree{} videos have gained considerable attention in recent years. However, challenges also arise in the streaming of such videos: 360\degree{} video sizes are much larger ($4\times$ to $6\times$) compared with conventional videos given the same perceived quality~\cite{qian2018flare}, due to their panoramic nature. The streaming is further challenged by the fact that streaming devices often have limited resources such as computational power and battery, and they are usually operated in wireless environments where bandwidth is scarce. 
This is especially relevant for the new generation of VR services and applications as envisioned by Metaverse. 

To reduce the cost of streaming 360\degree{} videos, one of the approaches that researchers have extensively studied is \textit{viewport prediction}. Generally, viewport prediction utilizes the previous viewing directions and head movements of one user\footnote{We use the terms ``user'' and ``video watcher'' interchangeably.} as knowledge to predict the user's viewport in the future. Most of the modern prediction approaches leverage machine learning (ML) as their primary tool, where the models include convolutional neural networks (CNNs)~\cite{zou2019probabilistic}, recurrent neural networks (RNNs)~\cite{fan2019optimizing}, long short-term memory (LSTM) networks~\cite{xu2018gaze}, etc. By predicting the viewport in the future, streaming devices can prefetch and cache video tiles around the predicted viewport center in advance to the HMDs. The requested video tiles are usually of high quality in the predicted viewport, and of low quality outside the predicted viewport (e.g., in~\cite{mahzari2018fov}). If during playback, the video tiles required in the current viewport are not present, the missing tiles will be downloaded from the server~\cite{song2019fast} to avoid playback stalls.

Despite the plethora of researches in this field, their main objectives are similar: improve the accuracy of the predicted viewport, so as to reduce the resources needed during playback or improve the quality of service (QoS). 
The primary goal in this paper, however, is not to design more accurate prediction algorithms. 
Instead, let us consider the following example where a group of people is watching the same video, and a single ML model is used for viewport prediction. This single ML model is trained based on past users' head movements and viewing directions while watching this video, and thus, it learns to represent a general watcher's behavior of this video. However, this single ML model may not be able to sufficiently serve a new video or a new user. The reasons are twofold: 1) Video contents are complex and diverse. Models trained offline with other videos' information may not be applicable to \textit{new} videos, e.g., live video streaming where there is no live analysis for region of interest (ROI); 2) User behaviors can be drastically different across users, and they can evolve over time. For instance, a user who is new to 360\degree{} videos may want to browse everywhere he or she can in a random manner at the beginning, while an experienced watcher may only focus on the object of interest all the time. In the first case, using a single ML model may result in poor prediction performance. Bearing this in mind, we focus on a related but different aspect of the viewport prediction problem: \textit{robustness}. That is, \textit{designing an adaptive and personalized viewport prediction paradigm to improve the worst-performing viewport prediction among users, while being resource-efficient}.

Looking into the literature, existing 360\degree{} video streaming systems usually assume there are a few pre-trained viewport prediction models ready for use, and assign the pre-trained model to each user for prediction~\cite{xie2018cls,qian2018flare, zhang2019drl360}. We argue that if the number of pre-trained models is too few, it is hard to account for the variety of user behaviors and video complexity. On the other hand, if there are too many pre-trained models for use, the users have to store all those models locally and constantly switch between them, which incurs excessive cost.

In light of the discussion, in this paper, we propose a novel streaming paradigm that bridges the gap between the literature and our goal. In the proposed paradigm, we adopt the idea from meta learning~\cite{thrun2012learning}, where a \textit{meta model} is trained according to a certain procedure (e.g. \cite{finn2017model}) with previous users' viewing behaviors. This training procedure is designed so that the resulting meta model is sensitive to new training data and can be adapted to a new training task with only a few training examples. The meta model, with a certain choice of design, can also serve as a good global model for the prediction of viewing directions (see Section~\ref{sec:robust_vp}). It is distributed to all users, and while each user plays the video, we utilize the initial short amount of user behaviors as our training examples to adapt the meta model to the specific user.

Besides improving the \textit{accuracy} of the prediction, there is another factor to consider: the \textit{size} of the viewport to prefetch. Intuitively, in order to have the entire actual viewport included in the prefetched video area, we can use a small (large) prefetch area for predictable (unpredictable) users, respectively. We use a second meta learning approach for the prefetch size. Similarly, we train a meta model for the prediction of the smallest viewport size that can include the actual viewport. Then, we adapt this meta model to each user.

The advantages of the \textit{double-meta-learning} paradigm can be summarized as follows.

\begin{itemize}
    \item \textbf{Robust viewport prediction.} We use \textit{two} meta learning based modules to ensure the robustness of viewport prediction: one for the viewing direction and the other for the prefetch size, where existing literature usually focus on the first module only. Those modules are fully personalized and adaptive for all types of viewing patterns. 
    \item \textbf{Resource efficiency.} The two meta models can be quickly adapted with only a few training examples. The adaptation is local, so there is no extra bandwidth incurred. We only need to keep two meta models for each video in the HMDs, which reduces storage usage.
    \item \textbf{Compatibility.} The model-agnostic meta learning (MAML)~\cite{finn2017model} that we use is compatible with a wide range of machine learning approaches (e.g., supervised regression and classification, reinforcement learning), making our algorithm also compatible to a variety of viewport prediction models.
\end{itemize}

To the best of our knowledge, we propose the first paradigm that aims to guarantee robustness in viewport prediction and improve the worst prediction performance. 
Evaluation results using both simulations and a system prototype demonstrate that our approach can improve the average successful video prefetch rate by up to 39\% for a group of users watching the same video, and by as much as 97\% for a specific user watching a video.
%We note that the source code of this paper will be made open-source upon paper acceptance to ensure the reproducibility of our results and facilitate future research in this area.

\textit{Roadmap:} The rest of the paper is organized as follows: Section \ref{sec:sys_model} describes the system model and introduces the notations. We propose three evaluation metrics in Section~\ref{sec:perf_metric}. Section~\ref{sec:robust_vp} and \ref{sec:simulations} present our meta learning algorithms and experiment results, respectively. We discuss related works in Section \ref{sec:related_work} and conclude our work in Section \ref{sec:conclusion}.

\begin{table}%\tablefootnote{Test footnote table}
  \caption{List of Main Notations$^*$}
  \label{tab:notations}
  \begin{tabular}{c||c}
    \hline
    Notations & Definition\\
    \hline
    $A(\cdot)$ & Area on a sphere\\
    $k$ &  (Local) SGD steps in meta learning\\
    $t, T$ & Discretized time, total time steps\\
    $\mathbf{u}(t)$ & The actual viewing direction at time $t$\\
    $\mathbf{v}(t)$ & The predicted viewing direction at time $t$\\
    $\mathcal{C}_{\cdot,\cdot}$ & A circle of a sphere defined by two params\\
    $N^\text{(d)}$, $N^\text{(a)}$ & \#Sampled tasks in the Reptile algorithms\\
    $\mathcal{M}_\cdot^\text{(d)}$, $\mathcal{M}_\cdot^\text{(a)}$ & An ML model subscripted by params\\
    $\mathcal{T}_i^\text{(d)}$, $\mathcal{T}_i^\text{(a)}$ & A task indexed $i$\\
    $\mathcal{T}^\text{(d)}$, $\mathcal{T}^\text{(a)}$ & $\mathcal{T}^\text{(d)}=\big\{\mathcal{T}_i^\text{(d)},\forall i\big\}$, $\mathcal{T}^\text{(a)}=\big\{\mathcal{T}_i^\text{(a)},\forall i\big\}$\\
    $\mathcal{L}_\cdot$ & Loss function w.r.t. the subscripted task\\
    $\mathcal{U}(\cdot)$ & Uniform distribution\\
    $\alpha$ & The fixed angle of the actual viewport\\
    $\beta(t)$ & The angle for prefetched viewport at time $t$\\
    $\gamma(t)$ & The observed angle between $\mathbf{u}(t)$ and $\mathbf{v}(t)$\\
    $\hat{\gamma}(t+1)$ & The predicted angle for time $t+1$\\
    $\eta^\text{(d)}$, $\eta^\text{(a)}$ & (Local) SGD learning rates\\
    $\epsilon^\text{(d)}$, $\epsilon^\text{(a)}$ & (Meta) SGD learning rates\\
    $\mu^\text{(d)}$, $\mu^\text{(a)}$ & SGD learning rates in the adaptation stage\\
    $\theta$ & Initial params for all VD models\\
    $\theta'_i$ & Adapted params for the $i$-th VD model\\
    $\phi$ & Initial params for all PA models\\
    $\phi'_i$ & Adapted params for the $i$-th PA model\\
  \hline
  \multicolumn{2}{l}{\footnotesize \begin{tabular}[l]{@{}l@{}} $^*$ Superscripts $\square^\text{(d)}$ and $\square^\text{(a)}$ always denote the viewing direction (VD)\\and prefetch angle (PA) tasks, respectively.\end{tabular}} \\
\end{tabular}
\end{table}

\section{System Model}\label{sec:sys_model}
In this section, we describe our modeling of viewport prediction and 360\degree{} video streaming. A list of main notations in this paper can be found in Table~\ref{tab:notations}.

\subsection{Viewport Prediction Model}

Let us consider a scenario where a user watches a 360\degree{} video. The video's panorama is modeled as a sphere that has a unit radius, whose center is located at a three-dimensional (3D) Euclidean coordinate system's origin. The watcher, residing at the sphere's center, watches the video in a certain direction described by a normalized 3D vector $\mathbf{u}(t)$. Time, denoted by $t$, is discretized to equal lengths. For the analysis in this paper, we make the following idealized assumptions.

\begin{itemize}
    \item The video frames can be arbitrarily split into infinitesimal regions, i.e., the video ``tiles'' can be considered as infinitesimal square blocks.
    \item The videos can be chunked into consecutive clips with arbitrary lengths.
    \item The actual viewport at any time is a \textit{circle of the sphere} with a fixed angle for ease of analysis. The choice of the shape can be easily extended to any realistic ones.
\end{itemize}
Those simplifications can be easily extended to real-world scenarios, and we implement a realistic emulation system in Appendix~\ref{sec:sys_prototype}.

Under those assumptions, the viewport can be depicted by two quantities: 1) one variable $\mathbf{u}(t)$ and 2) a fixed spanning angle $\alpha$ that depicts the size of the circle of the sphere. Similarly, the prefetched video at time $t$ can be depicted by two variables: 1) the predicted viewing direction $\mathbf{v}(t)$, which is also a normalized 3D vector, and 2) the spanning angle $\beta(t)$, and the resulting angle between $\mathbf{u}(t)$ and $\mathbf{v}(t)$ is denoted by $\gamma(t)$. Note that, unlike $\alpha$, $\beta(t)$ is a decision variable that can vary over time.

\begin{figure}
    \centering
    \includegraphics[width=0.9\linewidth]{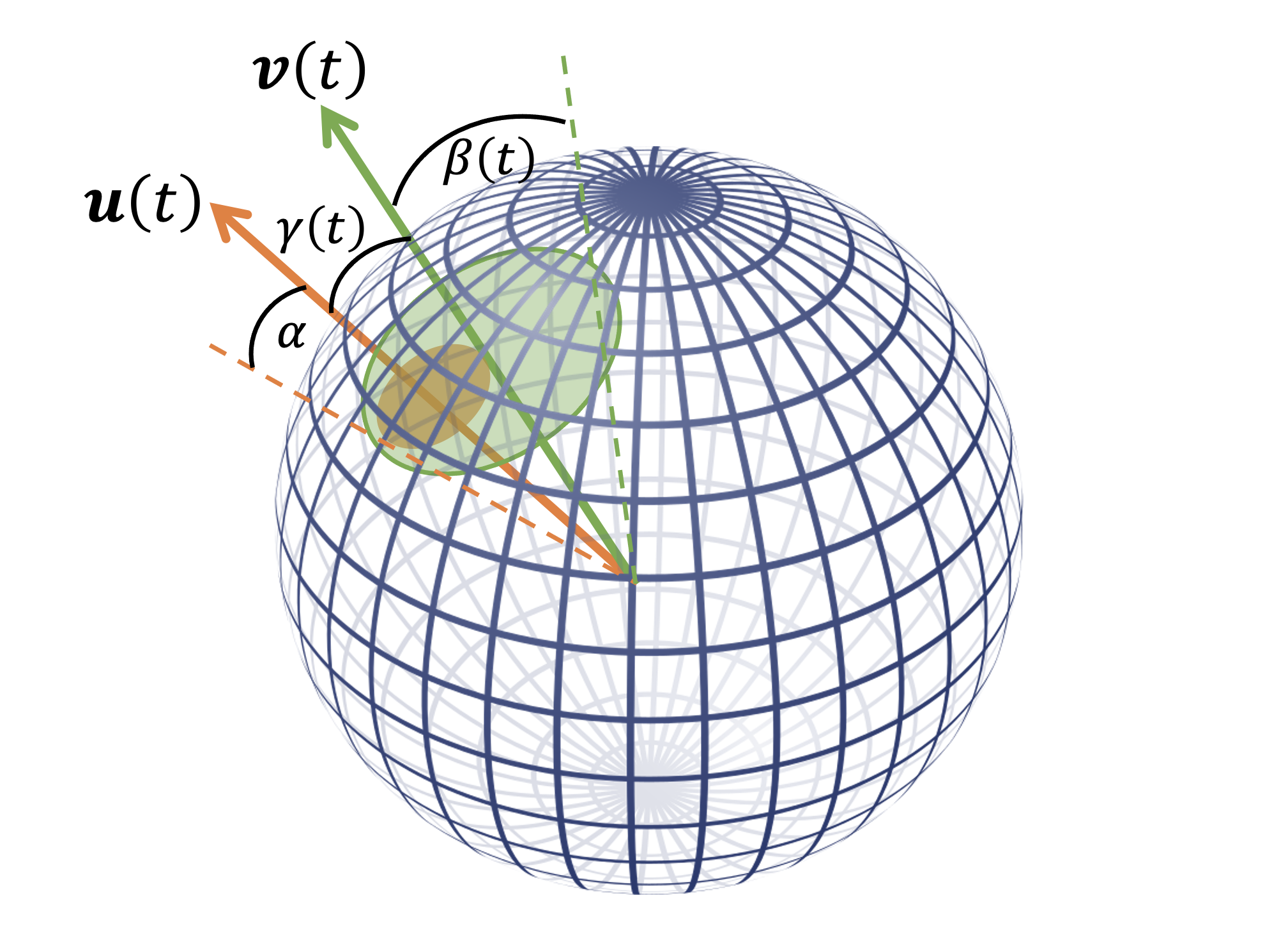}
    \caption{Illustration of viewport prediction.}
    \label{fig:vp_model}
\end{figure}

The illustration of our system model is presented in Figure~\ref{fig:vp_model}, where $\mathbf{u}(t)$ and $\mathbf{v}(t)$ are prolonged for ease of presentation.

\subsection{360\degree{} Video Streaming Model}\label{sec:video_streaming_model}
We assume a simplified video streaming system where at time $t$ the following is done in sequence for each user.
\begin{enumerate}
    \item The user calculates missing video tile indices in the current viewport.
    \item The user predicts $\mathbf{v}(t+1)$, decides $\beta(t+1)$ for time $t+1$, and calculates the video tiles to prefetch.
    \item The user requests both missing video tiles and the prefetch video tiles from the server.
    \item The server sends back the requested data to the user.
\end{enumerate}
%We depict our video streaming model in Figure~\ref{fig:stream_illus}.

\section{Performance Metrics}\label{sec:perf_metric}
To measure the accuracy of the 
viewport prediction, we propose three different performance metrics: Mean Angular Error (MAE), Mean Successful Prefetch Ratio (MSPR), and Improvement on the Worst Prediction (IWP).

\paragraph{Mean Angular Error} The MAE measures the average angular deviation between the predicted and the actual viewing direction (lower MAE values are desired).

\begin{equation}
    \text{MAE} = \frac{1}{T} \sum_{t=1}^T \gamma(t) =
    \frac{1}{T} \sum_{t=1}^{T} \cos^{-1} \left( \mathbf{u}(t) \cdot \mathbf{v}(t) \right) \,\,,
\end{equation}
where $T$ denotes the total time steps, $\mathbf{u}(t) \cdot \mathbf{v}(t)$ denotes the dot product between $\mathbf{u}(t)$ and $\mathbf{v}(t)$.

\paragraph{Mean Successful Prefetch Ratio} To precisely characterize how successful the prefetch is, we incorporate the decision of $\beta(t)$, and propose the second performance metric MSPR.
\begin{equation}
    \text{MSPR}=\frac{1}{T}\sum_{t=1}^{T} \frac{A\big(\mathcal{C}_{\mathbf{u}(t), \alpha} \cap \mathcal{C}_{\mathbf{v}(t), \beta(t)}\big)}{A\big(\mathcal{C}_{\mathbf{u}(t), \alpha}\big)}\,\,,
\end{equation}
where $A(\cdot)$ denotes the area (on a sphere), and $\mathcal{C}_{\cdot, \cdot}$ denotes the circle of a unit-radius sphere defined by the direction and the angle in the subscript. This performance metric characterizes the average fraction of successful prefetch\footnote{In practice, we use approximate integration methods to compute the intersection area which can be computationally costly, but this metric is for evaluation purposes only. It is not implemented on-device in real production.} (higher MSPR values are desired). Intuitively, the overlap between the green and orange areas in Figure~\ref{fig:vp_model} is considered successfully prefetched.

\paragraph{Improvement on the Worst Prediction} The primary goal of our proposed approach is to improve the robustness of the viewport prediction by alleviating the poor predictions. By slightly abusing the MSPR notation, we define the IWP metric as follows.

\begin{equation}
    \text{IWP}= \Delta \text{MSPR}_{i^*}\,\,,
\end{equation}
where $i^*:=\operatorname*{argmin}_i \text{MSPR}_i$ is the worst-performing index of tasks among all users w.r.t. a specific video, and $\Delta \text{MSPR}_{i^*}$ denotes the difference of the MSPR metric between our approach and the original method (higher IWP values are desired). On the other hand, improving the IWP should not hurt the performance of well-performing tasks at the same time. This will be investigated in Section~\ref{sec:simulations}.

In addition, we use auxiliary metrics such as the number of tiles transmitted from the server to the user in the evaluations.

% \begin{figure}[t]
%     \centering
%     \includegraphics[width=0.8\linewidth]{figs/illus_streaming.png}
%     \caption{Illustration of video streaming.}
%     \label{fig:stream_illus}
% \end{figure}

\section{Meta Learning Based Robust Viewport Prediction Paradigm}\label{sec:robust_vp}
In this section, we propose our paradigm. Generally speaking, the two decision variables $\mathbf{v}(t)$ and $\beta(t)$ can produce better performance metrics when $\mathbf{v}(t)$ is closer to $\mathbf{u}(t)$, and $\beta(t)$ is larger (but not too large so that the bandwidth is wasted). Both directions are explained in Section~\ref{sec:viewing_dir} and Section~\ref{sec:prefetch_size}, respectively. The overall algorithm is summarized in Section~\ref{sec:overall_alg}.

Note that we try to improve the metrics through the decision of $\mathbf{v}(t)$ and $\beta(t)$ only. Orthogonal elements such as tiling, encoding, saliency, etc. are compatible with our approach and are consequently out of our discussion in this paper.

\subsection{Prediction of the Viewing Direction}\label{sec:viewing_dir}

\subsubsection{Training the Meta Model}
Meta learning excels in learning 
 a new task quickly from a variety of tasks previously seen. 
 In this section, the task is to predict the viewing direction  $\mathcal{T}_i^\text{(d)}$, associated with video index $i$. 
 The input we can use in this task includes the sequence of visual saliency, camera rotations, etc. until the current time $t$, and the output is a prediction of the viewing direction $\mathbf{v}(t+1)$ for the next time slot. To accomplish the task, we first train a global meta model $\mathcal{M}_\theta^\text{(d)}$, parameterized by $\theta$, and then adapt $\mathcal{M}_\theta^\text{(d)}$ to each task $\mathcal{T}_i^\text{(d)}$. Here, the superscript denotes the direction prediction task.

Ideally, $\mathcal{M}_\theta^\text{(d)}$ is trained to be sensitive to unseen data, so that it can be adapted quickly. Assuming the model is adapted using stochastic gradient descent (SGD) for $k$ steps, we denote by $\theta'_i$ the parameter adapted to task $\mathcal{T}_i$, i.e.,
\begin{equation}\label{eq:sgd}
\theta'_i = \textup{SGD}(\mathcal{L}_{\mathcal{T}_i^\text{(d)}}, \theta, k, \eta^\text{(d)})\,\, ,
\end{equation}
where $\mathcal{L}_{\mathcal{T}_i^\text{(d)}}$ is the loss function w.r.t. task $\mathcal{T}_i^\text{(d)}$, and $\eta^\text{(d)}$ is the learning rate (for the local steps) within each task. In the MAML framework~\cite{finn2017model}, the meta model parameters $\theta$ are trained via optimizing $\mathcal{M}_{\theta'_i}^\text{(d)}$, such that $k$ steps of training on a new task will produce maximal effect. Assuming the tasks $\mathcal{T}_i^\text{(d)}$, $\forall i$ are of equal importance, and are sampled uniformly, our meta learning goal is:

\begin{equation}\label{eq:maml_goal}
\min_\theta \sum_{\mathcal{T}_i^\text{(d)} \sim \mathcal{U}(\mathcal{T}^\text{(d)})}\mathcal{L}_{\mathcal{T}_i^\text{(d)}} \left(\mathcal{M}_{\theta'_i} \right) \,\,,
\end{equation}
where $\mathcal{U}(\mathcal{T}^\text{(d)})$ denotes the uniform distribution across all viewing direction prediction tasks. To produce the maximum effect in one step, we select $k=1$ local step. It follows that (\ref{eq:sgd}) becomes: 
\begin{equation}\label{eq:sgd_one_step}
    \theta'_i = \theta -\eta^\text{(d)}\nabla_\theta \mathcal{L}_{\mathcal{T}_i^\text{(d)}}\left(\mathcal{M}_\theta^\text{(d)} \right)\,\,,
\end{equation}
and (\ref{eq:maml_goal}) becomes: 
\begin{equation}\label{eq:maml_goal_one_step}
\min_\theta \sum_{\mathcal{T}_i^\text{(d)} \sim \mathcal{U}\left(\mathcal{T}^\text{(d)}\right)} \mathcal{L}_{\mathcal{T}_i^\text{(d)}} \left(\mathcal{M}_{\theta-\eta^\text{(d)}\nabla_\theta \mathcal{L}_{\mathcal{T}_i^\text{(d)}}\left(\mathcal{M}_\theta^\text{(d)} \right)}^\text{(d)} \right) .
\end{equation}

\begin{algorithm}[t]
\caption{MAML for Viewing Direction Prediction}
\label{alg:maml_vd}
% \SetKwInOut{require}{Require}
% \require{}
Randomly initialize $\theta$;\\
\ForEach{\textup{iteration}}{
Sample viewing direction prediction tasks $\big\{\mathcal{T}_i^\text{(d)} \big\}$ uniformly;\\
\ForEach{$\mathcal{T}_i^\text{(d)}$}{
Sample training examples $\mathcal{D}$ from $\mathcal{T}_i^\text{(d)}$;\\
Evaluate $\nabla_\theta\mathcal{L}_{\mathcal{T}_i^\text{(d)}}(\mathcal{M}_\theta^\text{(d)})$ using $\mathcal{D}$ and $\mathcal{L}_{\mathcal{T}_i^\text{(d)}}$ according to~(\ref{eq:sgd_one_step});\\
Sample datapoints $\mathcal{D}'_i$ from $\mathcal{T}_i^\text{(d)}$ for the meta update;\\
}
Update the meta model using each $\mathcal{D}'_i$ and $\mathcal{L}_{\mathcal{T}_i^\text{(d)}}$ according to (\ref{eq:sgd_meta_update_vd}).
}
\end{algorithm}

\begin{algorithm}[t]
\caption{Reptile for Viewing Direction Prediction}
\label{alg:reptile_vd}
Randomly initialize $\theta$;\\
\ForEach{\textup{iteration}}{
Sample $N^\text{(d)}$ viewing direction prediction tasks $\big\{\mathcal{T}_i^\text{(d)}\big\}$ uniformly;\\
\ForEach{$\mathcal{T}_i^\text{(d)}$}{
Sample training examples $\mathcal{D}$ from $\mathcal{T}_i^\text{(d)}$;\\
Evaluate $\nabla_\theta\mathcal{L}_{\mathcal{T}_i^\text{(d)}}(\mathcal{M}_\theta^\text{(d)})$ using $\mathcal{D}$ and $\mathcal{L}_{\mathcal{T}_i^\text{(d)}}$, and obtain $\theta'_i$ according to~(\ref{eq:sgd_one_step});\\
}
Update $\theta$ according to (\ref{eq:meta_reptile_vd}).}
\end{algorithm}

To solve (\ref{eq:maml_goal_one_step}), we can use a meta SGD, one (meta) training step of which can be described by the following equation: 
\begin{equation}\label{eq:sgd_meta_update_vd}
    \theta \leftarrow \theta - \epsilon^\text{(d)} \nabla_\theta \sum_{\mathcal{T}_i^\text{(d)} \sim \mathcal{U}(\mathcal{T^\text{(d)}})} \mathcal{L}_{\mathcal{T}_i^\text{(d)}}\left(\mathcal{M}_{\theta'_i}^\text{(d)} \right) \,\,,
\end{equation}
where $\epsilon^\text{(d)}$ is the meta learning rate.
This MAML based algorithm is shown in Algorithm~\ref{alg:maml_vd}.

Computationally, the training of (\ref{eq:sgd_meta_update_vd}) involves a second-order optimization, which is expensive. To reduce the computational burden, we can be approximate it by a first-order optimization by omitting the second-order derivatives. We adopt a variation of the first-order MAML, namely, Reptile~\cite{nichol2018first} (Algorithm~\ref{alg:reptile_vd}). This remarkably simple algorithm is closely related to Algorithm~\ref{alg:maml_vd}, whose meta update is: 

\begin{equation}\label{eq:meta_reptile_vd}
\theta \leftarrow \theta - \epsilon^\text{(d)} \Delta \theta \,\,,
\end{equation}
where
\begin{equation}\label{eq:delta_theta}
    \Delta \theta := \frac{1}{N^\text{(d)}} \sum_{i=1}^{N^\text{(d)}} (\theta-\theta'_i) \,\,.
\end{equation}
In (\ref{eq:delta_theta}), $N^\text{(d)}:=\big\vert \big\{\mathcal{T}_i^\text{(d)}\big\} \big\vert$ is the number of sampled tasks in each iteration. Moreover, for our selection $k=1$, the expected $\Delta \theta$ (equivalent to $\theta$'s gradient) is: 

\begin{align}\label{eq:expectation_vd}
    \mathbb{E} \left[ \Delta \theta \right] &= \eta^\text{(d)} \cdot \sum_{\mathcal{T}_i^\text{(d)} \sim \mathcal{U}(\mathcal{T^\text{(d)}})} \nabla_\theta \mathcal{L}_{\mathcal{T}_i^\text{(d)}}\left(\mathcal{M}_\theta^\text{(d)} \right) \nonumber \\
    &=\eta^\text{(d)} \cdot \mathbb{E}_{\mathcal{T}_i^\text{(d)} \sim \mathcal{U}(\mathcal{T^\text{(d)}})} \left[ \nabla_\theta \mathcal{L}_{\mathcal{T}_i^\text{(d)}}\left(\mathcal{M}_\theta^\text{(d)} \right) \right] \,\,.
\end{align}

Equation (\ref{eq:expectation_vd}) indicates that, by selecting $k=1$, in expectation, parameter's ``gradient'' in Reptile and the expected gradient only differ by a constant factor (other statistical indicators such as variance could be different). For this reason, this Reptile meta model can be used as a good prediction model while being a sensitive model for adaptation.

\subsubsection{Meta Model Adaptation} After the meta model is obtained, it can be adapted to a new user during the video playback. This can be performed with a fixed time interval. Note that the learning rate in the adaptation stage can be different from that in the meta model training stage, and we denote it by $\mu^\text{(d)}$.

\begin{figure*}[t]
    \centering
    \includegraphics[width=0.85\linewidth]{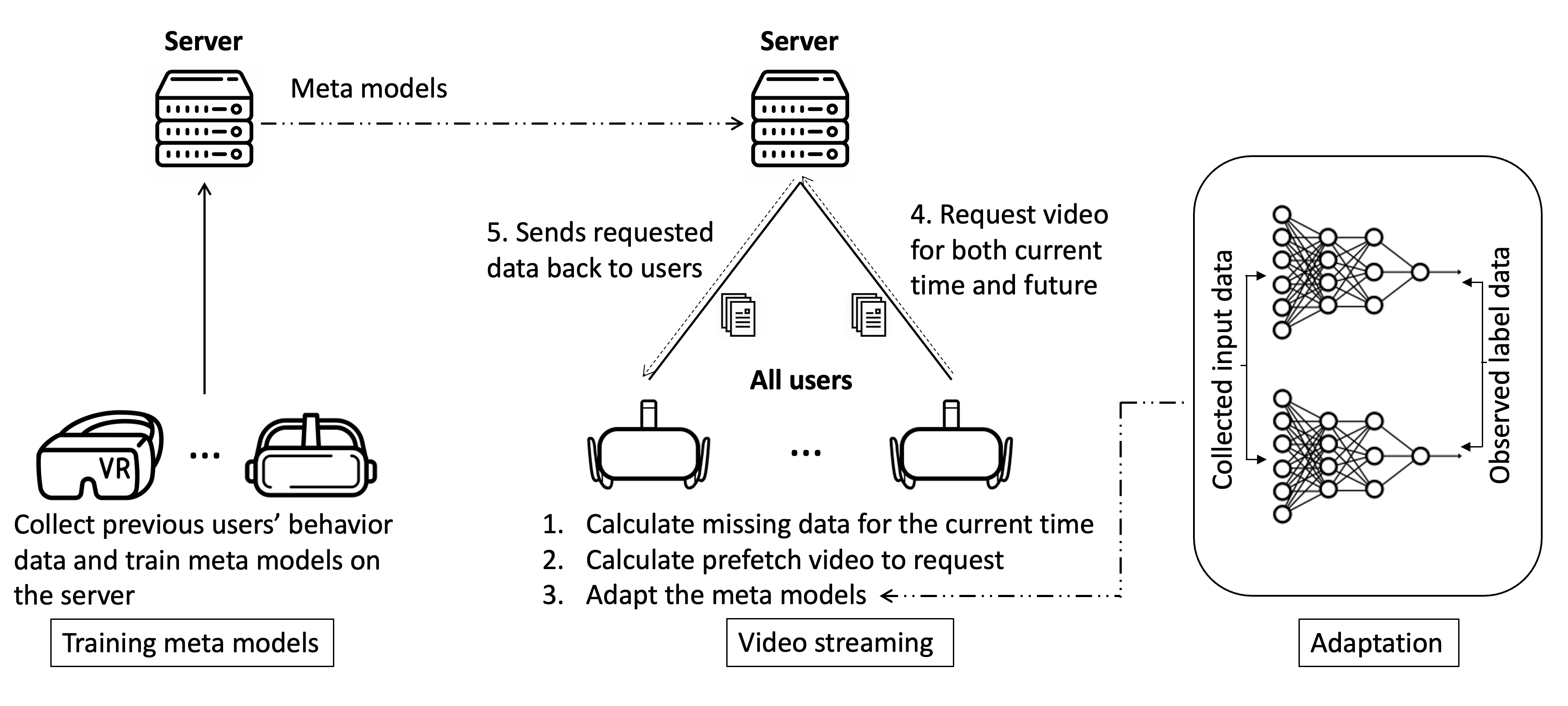}
    \caption{Illustration of the overall paradigm.}
    \label{fig:overall_illustration}
\end{figure*}

\subsection{Determination of the Prefetch Video Size}\label{sec:prefetch_size}

\subsubsection{Training the Meta Model}

If the streaming system's resources are unlimited, the prefetch size can be as large as possible. Realistically, however, prefetching too much can result in more battery consumption as well as more bandwidth usage that can compete with other users. On the other hand, prefetching insufficiently can result in suboptimal QoS. How to balance the trade-off is the topic of this section.

Similar to Section~\ref{sec:viewing_dir}, we leverage a meta learning based approach to solve this problem. Here, the input of task $i$, $\mathcal{T}_i^\text{(a)}$, is a sequence of $\gamma(t)$ collected during playback. The task objective is to predict $\hat{\gamma}(t+1)$ for the next time slot, which is a single value. The prefetch angle decision, $\beta(t+1)$, is the predicted minimum angle of the prefetched video that can cover the entire actual viewport.

\begin{equation}\label{eq:beta}
    \beta(t+1) = \hat{\gamma}(t+1)+\alpha\,\,.
\end{equation}

To train the meta model, we adopt the Reptile training schedule likewise. That is, we first train a sensitive meta model $\mathcal{M}_{\phi}^\text{(a)}$, parameterized by $\phi$, based on the past observed data. Then, the sequence of $\gamma(t)$ for a new video watcher is fed into the model for adaptation, and this model produces $\hat{\gamma}(t+1)$ over time. The observed $\gamma(t+1)$ values are used as labels to train $\mathcal{M}_{\phi}^\text{(a)}$. In the meta model training stage, one step of local update in the prefetch angle prediction is
\begin{equation}\label{eq:sgd_one_step_pa}
    \phi'_i = \phi -\eta^\text{(a)}\nabla_\phi \mathcal{L}_{\mathcal{T}_i^\text{(a)}}\left(\mathcal{M}_\phi^\text{(a)} \right)\,\,,
\end{equation}
where $\phi'_i$ denotes the adapted parameter for video index $i$. The meta update also becomes
\begin{equation}\label{eq:meta_reptile_pa}
\phi \leftarrow \phi - \frac{\epsilon^\text{(a)}}{N^\text{(a)}} \sum_{i=1}^{N^\text{(a)}} (\phi - \phi'_i) \,\,,
\end{equation}
where $N^\text{(a)} := \big\vert \big\{\mathcal{T}_i^\text{(a)}\big\} \big\vert$.
The update rule (\ref{eq:meta_reptile_pa}) shares the same property with (\ref{eq:meta_reptile_vd}). Thus, the meta model trained in this section also serves as a good prediction model for the viewport prefetch angle.

\begin{algorithm}[t]
\caption{Reptile for Viewport Prefetch Angle}
\label{alg:reptile_pa}
Randomly initialize $\phi$;\\
\ForEach{\textup{iteration}}{
Sample $N^\text{(a)}$ viewport angle prediction tasks $\big\{\mathcal{T}_i^\text{(a)}\big\}$ uniformly;\\
\ForEach{$\mathcal{T}_i^\text{(a)}$}{
Sample training examples $\mathcal{D}$ from $\mathcal{T}_i^\text{(a)}$;\\
Evaluate $\nabla_\phi\mathcal{L}_{\mathcal{T}_i^\text{(a)}}(\mathcal{M}_\phi^\text{(a)})$ using $\mathcal{D}$ and $\mathcal{L}_{\mathcal{T}_i^\text{(a)}}$, and obtain $\phi'_i$ according to~(\ref{eq:sgd_one_step_pa});\\
}
Update $\phi$ according to (\ref{eq:meta_reptile_pa}).
}
\end{algorithm}

\begin{algorithm}[t]
\caption{The Overall Paradigm}
\label{alg:robust_vp}
\tcp{Training meta models}
Randomly initialize $\theta$ and $\phi$ for meta models $\mathcal{M}_{\theta}^\text{(d)}$ and $\mathcal{M}_{\phi}^\text{(a)}$, respectively;\\
Train the meta model $\mathcal{M}_{\theta}^\text{(d)}$ following Algorithm~\ref{alg:reptile_vd};\\
Train the meta model $\mathcal{M}_{\phi}^\text{(a)}$ following Algorithm~\ref{alg:reptile_pa};\\
\ForEach{user, in parallel}{
Deploy $\mathcal{M}_{\theta}^\text{(d)}$ and $\mathcal{M}_{\phi}^\text{(a)}$ to the device;\\
\For{\textup{time} $t=1,2,\ldots$}{
\tcp{Calculate requested data}
Calculate the missing tiles in the current viewport;\\
Use outputs from $\mathcal{M}_{\theta}^\text{(d)}$ and $\mathcal{M}_{\phi}^\text{(a)}$ to calculate the video to prefetch;\\
Request both the missing tiles and the prefetch video from the server;\\
\tcp{Adapt meta models}
Collect training data locally;\\
Use collected input data and the actual viewing direction to adapt $\mathcal{M}_{\theta}^\text{(d)}$;\\
Use the error produced in $\mathcal{M}_{\theta}^\text{(d)}$ to adapt $\mathcal{M}_{\phi}^\text{(a)}$;\\
}
}
\end{algorithm}

Note that the prefetch angle model $\mathcal{M}_{\phi}^\text{(a)}$ is dependent on the performance of the viewing direction model $\mathcal{M}_{\theta}^\text{(d)}$: If the user is predictable, and $\mathcal{M}_{\theta}^\text{(d)}$ works accurately, $\mathcal{M}_{\phi}^\text{(a)}$ produces small prefetch angles. The training schedule resembles that in Section~\ref{sec:viewing_dir}, and it is described in Algorithm~\ref{alg:reptile_pa}. The superscript in this section means the angle prediction task.

\subsubsection{Meta Model Adaptation} The meta model adaptation for $\mathcal{M}_{\phi}^\text{(a)}$ is synchronized with the adaptation of $\mathcal{M}_{\theta}^\text{(d)}$. We denote the adaptation learning rate for $\mathcal{M}_{\phi}^\text{(a)}$ by $\mu^\text{(a)}$.

\subsection{The Overall Algorithm}\label{sec:overall_alg}
%Combining our system description along with the training and adaptation of two meta models, 
Our overall paradigm is roughly stated as follows. We first train the two meta models $\mathcal{M}_{\theta}^\text{(d)}$ and $\mathcal{M}_{\phi}^\text{(a)}$ based on past data following Algorithm~\ref{alg:reptile_vd} and Algorithm~\ref{alg:reptile_pa}, respectively. Then, we follow the video streaming model described in Section~\ref{sec:video_streaming_model}, with an extra step to adapt both $\mathcal{M}_{\theta}^\text{(d)}$ and $\mathcal{M}_{\phi}^\text{(a)}$ in each time step. The overall algorithm is stated in Algorithm~\ref{alg:robust_vp}, and the graphical illustration of our overall algorithm can be found in Figure~\ref{fig:overall_illustration}.

% Both the original version and the variations of our algorithm (e.g., different choices of $k$, full adaptation throughout playbacks versus adaptations for only the initial period of playbacks, etc.) are discussed in the evaluation sections.

\section{Experiments}\label{sec:simulations}

\subsection{Setup}
\subsubsection{Datasets} In the experiments, we use a dataset containing head-tracking data from 48 users watching 18 panoramic videos from 5 categories~\cite{wu2017dataset}. This dataset is collected in two separate experiments, aiming to serve two different purposes. The first experiment intends to capture the natural user behaviors when they are introduced to a new virtual environment. The second experiment investigates user behaviors in live VR streaming.

\subsubsection{Model Architectures} We use two separate LSTM networks~\cite{hochreiter1997long} for the viewing direction task and the prefetch angle task, i.e., $\mathcal{M}_{\theta}^\text{(d)}$ and $\mathcal{M}_{\phi}^\text{(a)}$, for the two prediction tasks. Note that the selection of models is not restrictive to LSTM networks. Specifically, the $\mathcal{M}_{\theta}^\text{(d)}$ model has 3D input and output, and one hidden layer of size 128. The $\mathcal{M}_{\phi}^\text{(a)}$ model has single-dimensional input and output, and one hidden layer of size 128.

The details of data pre-processing, training schedules, and hyperparameter selections are included in Appendix~\ref{sec:exp_setup}.

\subsubsection{Compatibility}\label{sec:compatibility}
It is worth mentioning that our algorithm is insensitive to the selection of setup, including ML models, sequence lengths, and other hyperparameters. For instance, replacing the LSTM networks with regular RNNs or even linear models will only result in inferior baselines, but we can still improve on the suboptimal models---the improvement could be smaller due to the incapacity of the selected models.

\subsubsection{Baselines}
Despite the fact that our algorithm is compatible with or orthogonal to the majority of viewport prediction algorithms, we compare our work with the CLS~\cite{xie2018cls} and CUB360~\cite{ban2018cub360} systems in the evaluations. In the CLS system, users' viewing directions are clustered into classes, and the probability of a tile being viewed by the user is calculated using the pre-trained model associated with the predicted user class. Simultaneously, there is an ML module that keeps predicting the user classes and is being trained by the observed user classes. In the CUB360 system, the prediction for one user leverages the predictions of this user's nearest neighbors in terms of viewing directions. The predictions of its $k$ nearest neighbors are combined with weighted average, where $k$ is a tunable parameter.

To strengthen the CLS system and make it comparable with our approach, we make the following alterations to it: 1) instead of predicting the probability of a tile being viewed, we predict the viewing direction; 2) we obtain pre-trained models for every user in every video, where the training schedule is the same as in Section~\ref{sec:training_schedule}, and 3) in every time step, instead of predicting the user class, we evaluate the prediction performance on all 48 models (corresponding to 48 user models in the same video), and we select the model that gives the best performance in the current time step. Note that our version of the CLS system's performance is \textit{an absolute upper bound of the original CLS system} since in 2), we use significantly more pre-trained models than the assumed in the CLS system, and in 3), we use the best model rather than the model associated with the predicted class, and most importantly, the selected best-performing model uses the \textit{future actual viewing direction} which is not available during playback. For this reason, we name the CLS baseline Enhanced-CLS (E-CLS). To make the CUB360 system comparable with our approach, we replace its linear regression component with the same LSTM network that we use, and we use the weighted average of the predicted viewing directions, instead of tiles. We call this variant CUB360-LSTM.

\begin{figure*}[t]
  \centering
  \includegraphics[width=0.9\linewidth]{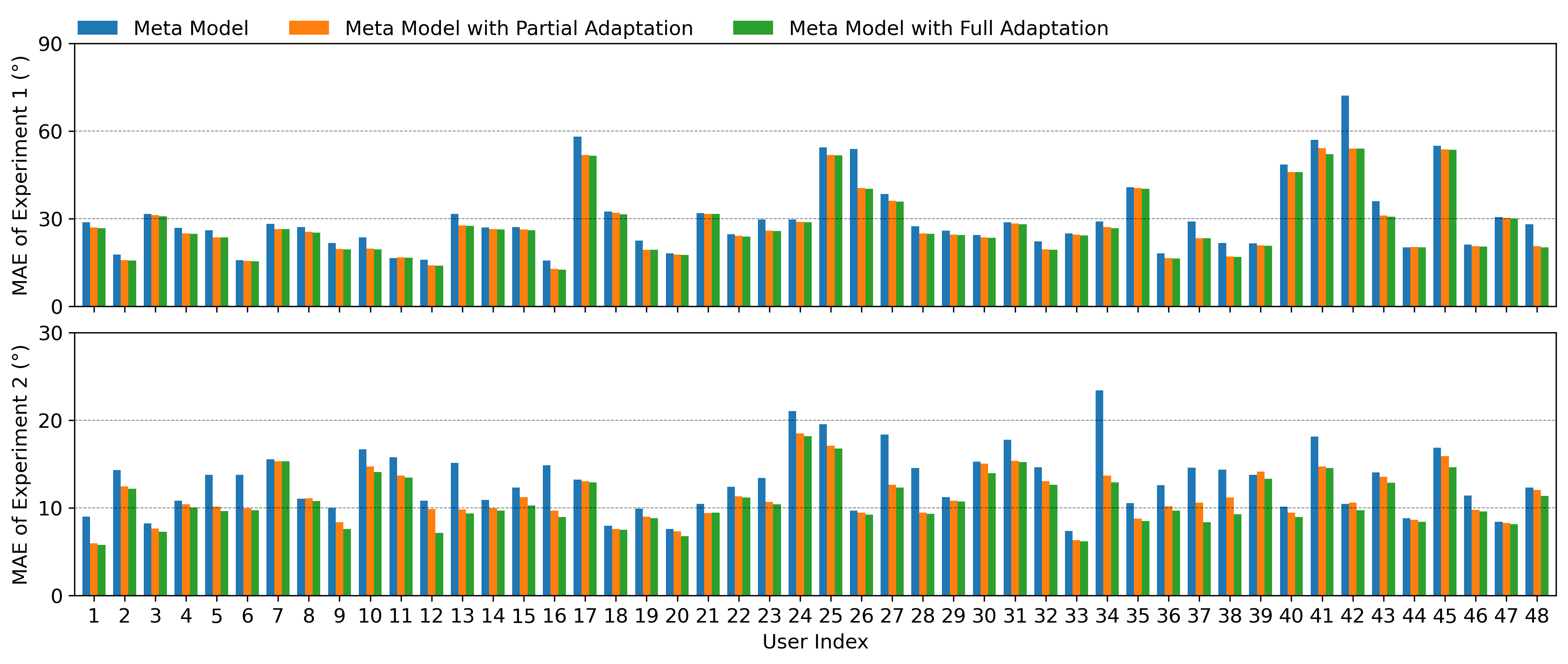}
  \caption{MAE metric of the first video in two experiments.}
  \label{fig:mae_first_video_two_exp}
\end{figure*}

\begin{figure*}[t]
  \centering
  \includegraphics[width=0.9\linewidth]{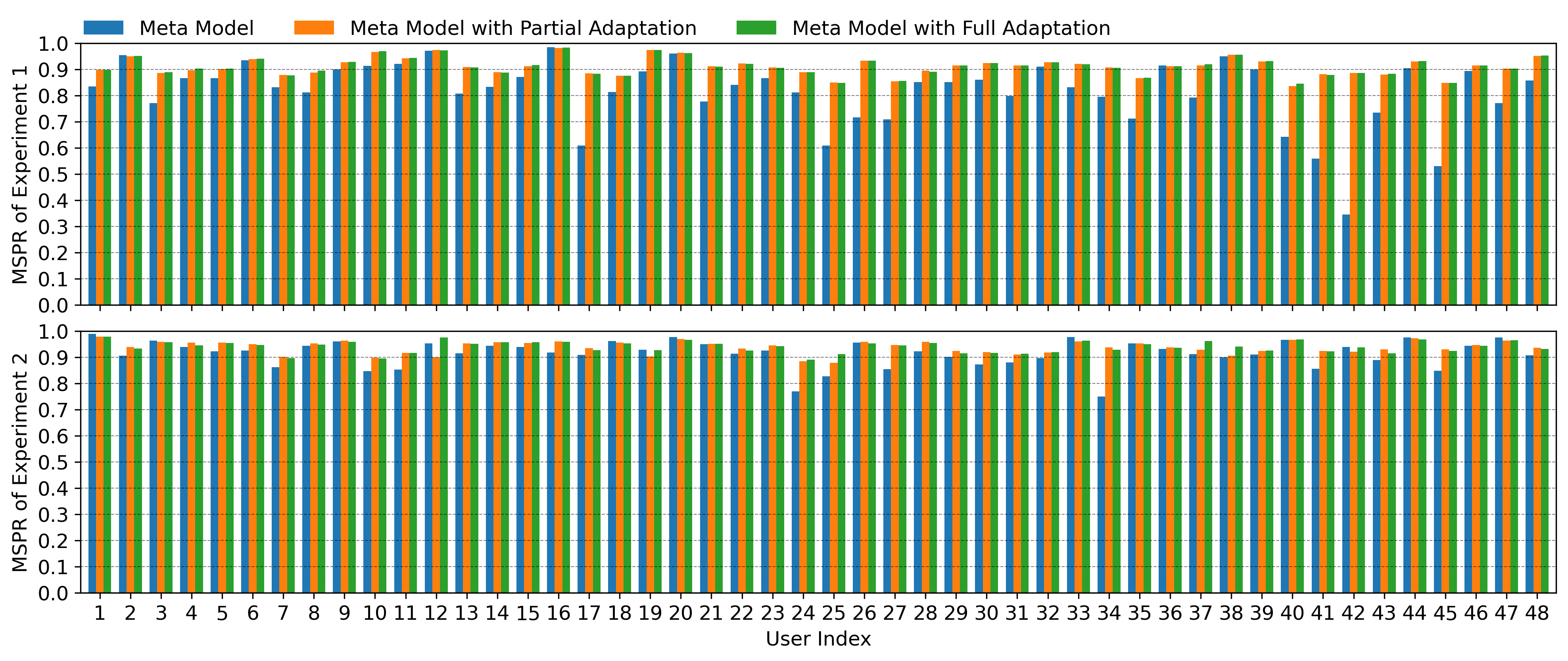}
  \caption{MSPR metric of the first video in two experiments.}
  \label{fig:mspr_first_video_two_exp}
\end{figure*}

\begin{table*}
  \caption{Performance Metrics of Two Experiments for Both Partial and Full Adaptation$^*$}
  \label{tab:perf_metrics}
  \footnotesize
  \begin{tabular}{ccccccccccc}
    \hline
    \begin{tabular}[c]{@{}c@{}}Exp.,\\Adpt.\end{tabular} & Metric & Video 1 & Video 2 & Video 3 & Video 4 & Video 5 & Video 6 & Video 7 & Video 8 & Video 9\\
    \hline
    \multirow{5}{*}{\begin{tabular}[c]{@{}c@{}}Exp. 1,\\Partial\end{tabular}} & $\Delta$MAE (\degree{}) & \begin{tabular}[c]{@{}c@{}}$[-18, 0]$,\\-3\end{tabular} & \begin{tabular}[c]{@{}c@{}}$[-24, 0]$,\\-6\end{tabular} & \begin{tabular}[c]{@{}c@{}}$[-41, 4]$,\\-7\end{tabular} & \begin{tabular}[c]{@{}c@{}}$[-24, 0]$,\\-3\end{tabular} & \begin{tabular}[c]{@{}c@{}}$[-46, 9]$,\\-14\end{tabular} & \begin{tabular}[c]{@{}c@{}}$[-32, 14]$,\\-8\end{tabular} & \begin{tabular}[c]{@{}c@{}}$[-58, 13]$,\\-10\end{tabular} & \begin{tabular}[c]{@{}c@{}}$[-17, -1]$,\\-5\end{tabular} & \begin{tabular}[c]{@{}c@{}}$[-9, 5]$,\\-1\end{tabular}\\
    \cline{2-11}
    
    & $\Delta$MSPR (\%) & \begin{tabular}[c]{@{}c@{}}$[0, 54]$,\\10\end{tabular} & \begin{tabular}[c]{@{}c@{}}$[0, 48]$,\\20\end{tabular} & \begin{tabular}[c]{@{}c@{}}$[17, 49]$,\\30\end{tabular} & \begin{tabular}[c]{@{}c@{}}$[-2, 41]$,\\8\end{tabular} & \begin{tabular}[c]{@{}c@{}}$[-8, 63]$,\\22\end{tabular} & \begin{tabular}[c]{@{}c@{}}$[2, 48]$,\\28\end{tabular} & \begin{tabular}[c]{@{}c@{}}$[-2, 76]$,\\36\end{tabular} & \begin{tabular}[c]{@{}c@{}}$[4, 40]$,\\17\end{tabular} & \begin{tabular}[c]{@{}c@{}}$[-11, 31]$,\\13\end{tabular}\\
    \cline{2-11}
    
    & IWP (\%) & 54 & 48 & 49 & 41 & 63 & 47 & 76 & 40 & 4\\
    \hline
    \multirow{5}{*}{\begin{tabular}[c]{@{}c@{}}Exp. 1,\\Full\end{tabular}} & $\Delta$MAE (\degree{}) & \begin{tabular}[c]{@{}c@{}}$[-18, 0]$,\\-3\end{tabular} & \begin{tabular}[c]{@{}c@{}}$[-34, -1]$,\\-9\end{tabular} & \begin{tabular}[c]{@{}c@{}}$[-45, -7]$,\\-17\end{tabular} & \begin{tabular}[c]{@{}c@{}}$[-24, 0]$,\\-3\end{tabular} & \begin{tabular}[c]{@{}c@{}}$[-49, 3]$,\\-16\end{tabular} & \begin{tabular}[c]{@{}c@{}}$[-40, -2]$,\\-19\end{tabular} & \begin{tabular}[c]{@{}c@{}}$[-63, -4]$,\\-18\end{tabular} & \begin{tabular}[c]{@{}c@{}}$[-17, -1]$,\\-5\end{tabular} & \begin{tabular}[c]{@{}c@{}}$[-33, -2]$,\\-10\end{tabular}\\
    \cline{2-11}
    
    & $\Delta$MSPR (\%) & \begin{tabular}[c]{@{}c@{}}$[0, 54]$,\\10\end{tabular} & \begin{tabular}[c]{@{}c@{}}$[0, 49]$,\\21\end{tabular} & \begin{tabular}[c]{@{}c@{}}$[17, 53]$,\\31\end{tabular} & \begin{tabular}[c]{@{}c@{}}$[-2, 42]$,\\8\end{tabular} & \begin{tabular}[c]{@{}c@{}}$[-6, 62]$,\\24\end{tabular} & \begin{tabular}[c]{@{}c@{}}$[5, 48]$,\\30\end{tabular} & \begin{tabular}[c]{@{}c@{}}$[7, 87]$,\\39\end{tabular} & \begin{tabular}[c]{@{}c@{}}$[4, 40]$,\\17\end{tabular} & \begin{tabular}[c]{@{}c@{}}$[1, 44]$,\\19\end{tabular}\\
    \cline{2-11}
    
    & IWP (\%) & 54 & 49 & 53 & 42 & 62 & 48 & 87 & 40 & 44\\
    \hline
    \multirow{5}{*}{\begin{tabular}[c]{@{}c@{}}Exp. 2,\\Partial\end{tabular}} & $\Delta$MAE (\degree{}) & \begin{tabular}[c]{@{}c@{}}$[-10, 0]$,\\-2\end{tabular} & \begin{tabular}[c]{@{}c@{}}$[-12, 7]$,\\-3\end{tabular} & \begin{tabular}[c]{@{}c@{}}$[-65, 0]$,\\-8\end{tabular} & \begin{tabular}[c]{@{}c@{}}$[-8, 1]$,\\-2\end{tabular} & \begin{tabular}[c]{@{}c@{}}$[-8, 0]$,\\-2\end{tabular} & \begin{tabular}[c]{@{}c@{}}$[-80, 0]$,\\-17\end{tabular} & \begin{tabular}[c]{@{}c@{}}$[-10, 4]$,\\-2\end{tabular} & \begin{tabular}[c]{@{}c@{}}$[-17, 2]$,\\-2\end{tabular} & \begin{tabular}[c]{@{}c@{}}$[-8, 2]$,\\-1\end{tabular}\\
    \cline{2-11}
    
    & $\Delta$MSPR (\%) & \begin{tabular}[c]{@{}c@{}}$[-5, 19]$,\\2\end{tabular} & \begin{tabular}[c]{@{}c@{}}$[-3, 40]$,\\6\end{tabular} & \begin{tabular}[c]{@{}c@{}}$[-1, 80]$,\\11\end{tabular} & \begin{tabular}[c]{@{}c@{}}$[-5, 20]$,\\4\end{tabular} & \begin{tabular}[c]{@{}c@{}}$[-3, 16]$,\\2\end{tabular} & \begin{tabular}[c]{@{}c@{}}$[0, 97]$,\\37\end{tabular} & \begin{tabular}[c]{@{}c@{}}$[-2, 19]$,\\4\end{tabular} & \begin{tabular}[c]{@{}c@{}}$[-2, 52]$,\\5\end{tabular} & \begin{tabular}[c]{@{}c@{}}$[-3, 13]$,\\3\end{tabular}\\
    \cline{2-11}
    
    & IWP (\%) & 19 & 40 & 80 & 13 & 15 & 97 & 3 & 52 & 6\\
    \hline
    \multirow{5}{*}{\begin{tabular}[c]{@{}c@{}}Exp. 2,\\Full\end{tabular}} & $\Delta$MAE (\degree{}) & \begin{tabular}[c]{@{}c@{}}$[-11, 0]$,\\-2\end{tabular} & \begin{tabular}[c]{@{}c@{}}$[-16, 0]$,\\-4\end{tabular} & \begin{tabular}[c]{@{}c@{}}$[-67, 0]$,\\-9\end{tabular} & \begin{tabular}[c]{@{}c@{}}$[-9, -1]$,\\-4\end{tabular} & \begin{tabular}[c]{@{}c@{}}$[-8, 0]$,\\-2\end{tabular} & \begin{tabular}[c]{@{}c@{}}$[-80, 0]$,\\-19\end{tabular} & \begin{tabular}[c]{@{}c@{}}$[-12, 0]$,\\-3\end{tabular} & \begin{tabular}[c]{@{}c@{}}$[-17, 0]$,\\-3\end{tabular} & \begin{tabular}[c]{@{}c@{}}$[-10, -1]$,\\-3\end{tabular}\\
    \cline{2-11}
    
    & $\Delta$MSPR (\%) & \begin{tabular}[c]{@{}c@{}}$[-1, 18]$,\\3\end{tabular} & \begin{tabular}[c]{@{}c@{}}$[-2, 37]$,\\7\end{tabular} & \begin{tabular}[c]{@{}c@{}}$[-1, 80]$,\\12\end{tabular} & \begin{tabular}[c]{@{}c@{}}$[-1, 18]$,\\6\end{tabular} & \begin{tabular}[c]{@{}c@{}}$[-2, 16]$,\\3\end{tabular} & \begin{tabular}[c]{@{}c@{}}$[1, 97]$,\\38\end{tabular} & \begin{tabular}[c]{@{}c@{}}$[-2, 21]$,\\5\end{tabular} & \begin{tabular}[c]{@{}c@{}}$[-2, 53]$,\\5\end{tabular} & \begin{tabular}[c]{@{}c@{}}$[-2, 13]$,\\4\end{tabular}\\
    \cline{2-11}
    
    & IWP (\%) & 18 & 37 & 80 & 15 & 16 & 97 & 17 & 53 & 11\\
    \hline
    \multicolumn{11}{l}{\footnotesize \begin{tabular}[l]{@{}l@{}}$^*$ ``Exp.'' stands for ``Experiment'', ``Adpt'' stands for ``Adaptation''. See definitions of ``partial'' and ``full'' in Section~\ref{sec:improve_vp}. For the $\Delta$MAE and $\Delta$MSPR\\metrics, the first value is a min-max range of metric change, and the second value is the average change, compared with the single global model\\approach. The units for MAE, MSPR and IWP metrics are degree, percentage (\%), and percentage (\%), respectively\end{tabular}} \\
\end{tabular}
\end{table*}

\subsection{Results}
\subsubsection{Improving the Viewport Prediction}\label{sec:improve_vp}
We validate the effectiveness of our algorithm through the observation of performance metrics' improvement and a comparison with the enhanced baseline approach CLS.
% \begin{figure*}[t]
%   \centering
%   \includegraphics[height=0.647cm]{figs/legend1.png}
% %   \caption{}
% \end{figure*}
% \begin{figure*}[t]
% \vspace{-5mm}
%   \centering
%   \includegraphics[width=\linewidth]{figs/mae_exp1_video1.png}
%   \caption{MAE}
% \end{figure*}

In Figure~\ref{fig:mae_first_video_two_exp} and Figure~\ref{fig:mspr_first_video_two_exp}, we present the comparison (for both experiments) of MAE and MSPR metrics, respectively. We plot the first video's results in both experiments, and the full results are in Table~\ref{tab:perf_metrics}. The comparisons are between 1) using a single global model, which is the meta model we train since we prove the equivalence for $k=1$ in Section~\ref{sec:viewing_dir},  2) using partial adaptation for the first 1,200 steps (120~s, for both meta models), and 3) using full adaptation for the entire video playback.

\begin{figure}[ht]
    \centering
    \includegraphics[width=1.\linewidth]{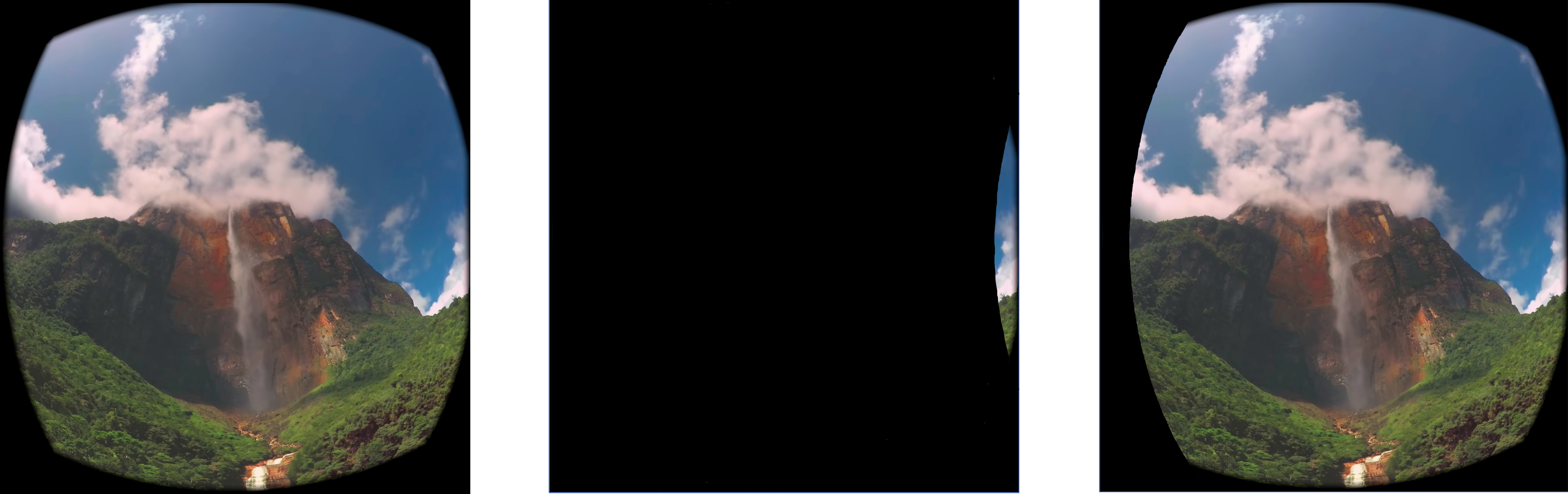}
    \caption{Visualization of prefetch ratios from a 360\degree{} video screenshot (single eye). Left: original screenshot; Middle: prefetch area with 85\degree{} angular error; Right: prefetch area with 5\degree{} angular error.}
    \label{fig:prefetch_vis}
\end{figure}

\paragraph{Comparing the MAE metric}
In Figure~\ref{fig:mae_first_video_two_exp}, we desire lower MAE values after (partial or full) adaptation, which is clearly shown in this bar plot. In experiment 1 (full adaptation), the maximum MAE improvement is 18\degree{}, while the worst adaptation only increases this metric by 0.14\degree{}. Among all users and all videos, the best improvement on the MAE metric is 80\degree{}. To visualize this substantial improvement, we assume that the prefetch area has the same shape as the viewport, and we plot a screenshot of a 360\degree{} video (from a single eye's view) in Figure~\ref{fig:prefetch_vis}. In this figure, the left shows the original screenshot; the middle and the right show the video regions where the colored areas are successfully prefetched for 85\degree{} and 5\degree{} angular errors, respectively, indicating the 80\degree{} difference in angular error.

\paragraph{Comparing the MSPR metric}
The MAE metric is solely determined by the viewing direction prediction module, while the MSPR metric involves decisions made by the prefetch size prediction module as well. For a fair comparison, we compare our algorithm with a single global model under the \textit{same bandwidth consumption}. We do so by first evaluating the bandwidth consumed by our algorithm, and evenly distribute the bandwidth to every prefetch with the single global model. 

It is apparent from Figure~\ref{fig:mspr_first_video_two_exp} that our algorithm (either partial or full adaptation) can increase the MSPR metric significantly. Note that the MSPR values are high in both experiments because our goal is to make sure the actual viewport always falls within the prefetched video. The prefetch size can be scaled down according to the network resource availability. This figure shows that, using the same amount of bandwidth, our algorithm can successfully prefetch more video. Or equivalently, with the same successful prefetch rate, our algorithm uses less bandwidth.

\paragraph{Results for the IWP metric}
As a result, the IWP metrics with full adaptation are 54\% and 18\% for experiment 1 and experiment 2, respectively. In experiment 1, our algorithm improved the worst successful prefetch rate from 35\% to 89\%. In experiment 2, the number increases from 75\% to about 93\%. The significant improvements on the worst-performing tasks validate our claim of robustness in viewport prediction.

The complete results of performance metrics are shown in Table~\ref{tab:perf_metrics}. For the MAE and MSPR results, we first present the range of metric change, and then the average metric change. Note that the MAE metric desires a negative change while the MSPR and IWP metrics desire positive changes. The results in Table~\ref{tab:perf_metrics} are in line with previous observations. Notably, we improve the MSPR by up to 39\% for a group of users watching the same video (video 7 in experiment 1), and by as much as 97\% for a specific user watching a video (video 6 in experiment 2) in the best case. This is a huge improvement which means that we adapt the pre-trained model so that a previously completely unpredictable user becomes almost predictable, leading to significantly improved QoS for this user.

We notice that the second experiment's results, in general, show a better performance compared with the first experiment. This is because the videos in the second experiments are live VR streaming, where the positions and the shooting directions of cameras are often fixed. Moreover, there is usually only one ROI in those videos, making the predictions easier~\cite{wu2017dataset}. We find our algorithm always improves the original method's performance regardless of the video types.

\paragraph{Comparison with the enhanced CLS and CUB360 baselines}
We compare the MAE and MSPR metrics between our approach and the baselines, where MSPR metrics are measured under the same bandwidth consumption. Only the full adaptation scenario is presented. We average the performance of all videos, users and time steps within one experiment. From Table~\ref{tab:compare_cls_avg}, we find that our approach apparently works better compared to the CUB360-LSTM approach. Also, our approach works comparably to the E-CLS system in terms of MAE, but because we have the prefetch angle module, our approach can significantly outperform the E-CLS system in terms of MSPR. This shows that, even comparing with the near-optimal version of the ensemble-based approach, ours is superior.

\begin{table}[t]
  \caption{Averaged Performance Metrics for Comparison with Baselines}
  \label{tab:compare_cls_avg}
  \normalsize
  \begin{tabular}{cccc}
    \hline
     Experiment & Approach & $\overline{\text{MAE}}$ (\degree{}) & $\overline{\text{MSPR}}$ (\%) \\
    \hline
     \multirow{3}{*}{Experiment 1} & E-CLS & 35 & 77 \\
     \cline{2-4}
     & CUB360-LSTM & 47 & 69 \\
     \cline{2-4}
     & \textbf{Ours} & 33 & 95 \\
     \hline
     \multirow{3}{*}{Experiment 2} & E-CLS & 16 & 91 \\
     \cline{2-4}
     & CUB360-LSTM & 24 & 85 \\
     \cline{2-4}
     & \textbf{Ours} & 19 &  93\\
    \hline
\end{tabular}
\end{table}

\subsubsection{Quick Adaptation}\label{sec:quick_adaptation}
As we discuss previously, two types of adaptation (partial and full) are used in the performance comparisons. In the dataset, the video length varies from 164 seconds to 655 seconds. Hence, the first 120 seconds make up 18\% to 73\% of the entire video length. From Table~\ref{tab:perf_metrics}, we show that with partial, or even a very short initial adaptation, our algorithm can still improve the original method to a level that is close to full adaptation.

\subsubsection{Resource Efficiency Analysis}
Our proposed paradigm is resource efficient because 1) regarding computation, we need only a few adaptation steps thanks to the meta models' sensitivity, and 2) regarding communication, we incur no extra cost. The details are as follows.

\paragraph{Extra computation cost}
We use the floating-point operations (FLOPs) as an indicator for our ML models' computational efficiency (see Appendix~\ref{appendix:flops}). Using the convention from the literature~\cite{tang2018flops}, we consider one addition or one multiplication operation as one FLOP. For the LSTM networks that we use, using existing conclusions from~\cite{zhang2018navigating}, a single-layer LSTM's FLOPs in forward pass is $2\times S \times (I + D) \times D \times 4$, where the factor $4$ accounts for the multiple gates in LSTM networks, and $S$, $I$, $D$ represent the sequence length, input, and hidden dimension, respectively. Taking the backward pass into account, numerically, the $\mathcal{M}_\theta^\text{(d)}$ model needs 40.2 MFLOPs per training cycle, and $\mathcal{M}_\phi^\text{(a)}$ model needs 39.6 MFLOPs per training cycle. Those numbers are linearly scaled with the total adaptation steps.

We claim that our system is lightweight for that the adaptation steps performed on the device can be much smaller than the total time steps $T$ (see Section~\ref{sec:quick_adaptation}).

For a rough estimate, the NVIDIA GeForce RTX 3080 GPU has 29.77 TFLOPs per second and requires 350 Watt power, which translates to $1.18\times10^{-11}$ joule per FLOP. The Oculus Quest 2 device has 14 Watt-hour (50,400 joules) of battery capacity, which means the battery can support

$$
\frac{50400}{1.18\times10^{-11} \times (40.2 + 39.6)\times10^6} = 5.35\times 10^7
$$
adaptation steps. With the training interval of 100~ms, the device's battery can support $5.35\times 10^6$ seconds (1,486 hours) duration, excluding all other sources that consume energy. Notice that in the experiments, we only use 1,200 steps (120 s). This energy consumption is negligible compared with the energy consumed by video playbacks.

\paragraph{Extra bandwidth consumption} Our framework essentially incurs no extra bandwidth consumption compared with conventional viewport prediction designs. This is because the adaptations are performed locally. Indeed, our algorithm may request larger viewports for less predictable users, but this is not considered an extra overhead introduced by the framework itself, and the comparisons are made fair in the evaluations.

\subsubsection{Additional Experimental Results} In Appendix Additional~\ref{sec:exp_other_k}, we present the results for another choice of $k$. We observe consistent results with results in the main paper.

\subsubsection{System Prototype}
We build a realistic emulation system prototype on the Oculus Quest 2 device, and test the system with 360\degree{} videos. The results are also consistent with the observations in the main paper. The details are in Appendix~\ref{sec:sys_prototype}.

\section{Related Work}\label{sec:related_work}

\subsubsection{Accurate Viewport Prediction} Researchers propose various approaches to improve the accuracy of viewport prediction in 360\degree{} videos. Traditional prediction methods include linear regression~\cite{bao2016shooting, petrangeli2017http}, probabilistic models~\cite{xie2017360probdash}, support vector regression~\cite{yang2019fovr}, as well as more complex statistical tools such as Gaussian filtering~\cite{feng2019viewport}. Deep neural networks such as CNNs~\cite{zou2019probabilistic} are applied for viewport prediction. Still, the main ML based instruments used in viewport prediction are RNNs~\cite{wu2020spherical} and their variants LSTM networks~\cite{fan2017fixation, zhang2019drl360}, due to their ability to deal with temporal dynamics. Cross-user behaviors are explored in~\cite{ban2018cub360,xie2018cls} to improve the prediction and reduce the variance in quality. Saliency is also utilized to help understand the ROI in a video~\cite{nguyen2018your, xu2018gaze}. For example, Nguyen \textit{et al.}~\cite{nguyen2018your} train a dedicated 360\degree{} saliency detection network based on human fixation on 360\degree{} videos, and predict users’ future viewport.

\subsubsection{Resource-efficient Viewport Prediction} Some other works focus on the resource efficiency in streaming 360\degree{} videos, most of which aiming at reducing the bandwidth requirement in streaming, for example, through tiling~\cite{corbillon2017viewport, graf2017towards,zhou2018clustile,hosseini2016adaptive}, encoding and decoding~\cite{sreedhar2016viewport}, bit-rate adaptation~\cite{spiteri2020bola}, and caching~\cite{mahzari2018fov}. Moreover, researches such as \cite{guan2019pano, zhang2019drl360} integrate multiple aforementioned aspects into a unified system design.

\subsubsection{Robust Viewport Prediction} Despite the rich literature that tries to improve the quality and resource efficiency of viewport prediction, there are few discussions in the literature that touch  upon the robustness of viewport prediction. In \cite{xie2018cls}, Xie \textit{et al.} propose a method that classifies users into groups based on their historical behavior, and predicts future viewport by groups to increase the accuracy. However, users are hard to be grouped into a few categories, so each user’s prediction model is not fully tailored. There are also approaches that improve robustness through aspects other than viewport prediction, such as video encoding and decoding~\cite{palash2021robust} and viewport prediction noise~\cite{guan2019pano}. Those works are orthogonal to ours, and they are out of the scope of our discussion.

\subsubsection{Meta Learning} Meta learning~\cite{thrun2012learning} aims to train a model on a variety of tasks such that this meta model can learn new tasks quickly using only a few training samples. Finn \textit{et al.} propose in~\cite{finn2017model} a model-agnostic meta learning method which is compatible with model representation that is amenable to gradient-based training and any differentiable objectives. We adopt a variation of the first-order approximation of MAML, Reptile~\cite{nichol2018first}, for more efficient implementation. Moreover, we design a pipeline of two such meta learning models tailored to the specific challenges of our problem.

\section{Conclusion}\label{sec:conclusion}
In this paper, we investigate a novel problem: the robustness in viewport prediction. Our objective is to design a resource-efficient paradigm that improves the performance of viewport prediction---especially for the worst-performing cases---through model adaptation. Our algorithm consists of two modules, both based on meta learning. The first direction prediction module adapts the meta model for a more accurate prediction of viewing direction, and the second prefetch angle module takes the prediction errors of viewing direction from the first module and decides how much to prefetch for enhanced robustness. Our approach is resource-efficient and it is compatible with a wide range of ML models. Simulations and emulations demonstrate that our algorithm can improve various performance metrics with negligible extra overhead. 

% \section*{Acknowledgment}

% The preferred spelling of the word ``acknowledgment'' in America is without 
% an ``e'' after the ``g''. Avoid the stilted expression ``one of us (R. B. 
% G.) thanks $\ldots$''. Instead, try ``R. B. G. thanks$\ldots$''. Put sponsor 
% acknowledgments in the unnumbered footnote on the first page.

\bibliographystyle{IEEEtran}
\bibliography{main}


\appendices

\section{Experiment Setup Details}\label{sec:exp_setup}
In this appendix, we provide details for the experiment setup.

\subsection{Data Pre-processing}
This dataset contains each user's time stamps, camera quaternions, and HMD's positions. We select the time stamps and transform the camera quaternions to 3D rotations as our training data. HMD positions are omitted because they do not have an effect on the viewport in the context of 360\degree{} videos. In the experiments, time is split evenly into 100~ms chunks. We average all data entries within every 100~ms time in the dataset\footnote{We use this setting despite the fact that some video formats cannot be chunked exactly to 100~ms chunks in reality.}.

% \subsection{Model Architectures}
% Specifically, the $\mathcal{M}_{\theta}^\text{(d)}$ model has 3D input and output, and one hidden layer of size 128. The $\mathcal{M}_{\phi}^\text{(a)}$ model has single-dimensional input and output, and one hidden layer of size 128.

\subsection{Training Schedules}\label{sec:training_schedule}
To train the $\mathcal{M}_{\theta}^\text{(d)}$ model, we take the previous 100 viewing direction data samples (not including the current one) as inputs and use the current viewing direction as the training label. Notice that the sequence length of 100 corresponds to a 10-second time period as the time interval is set to 100~ms. To train the $\mathcal{M}_{\phi}^\text{(a)}$, the inputs are the previous 100 $\gamma$ values, and the current observed $\gamma$ is used as the training label. The loss function we adopt is the mean squared loss. We clamp the decision $\beta(t+1)$ into the range of $[\alpha, \pi/2]$ in our simulations.

This schedule is used for both training the initial meta models and adapting the initial meta models. In the adaptation stage, we assume the first ten seconds of the video have already been stored locally on the device, so that we can start training and prediction from the 10\textsuperscript{th} second. We do not produce any evaluation data in the initial time that is equal to the sequence length.

We train the two meta models for each user w.r.t. each video. When doing so, the training data are selected from all other users' data from the same video, excluding this particular user's data (similar to leave-one-out cross-validation).

\subsection{Other Hyperparameters} We empirically choose the user's fixed viewport angle $\alpha=\pi/8$. Main hyperparameters are listed in Table~\ref{tab:hyperparams} for reference.

\section{Additional Experiment Results}\label{sec:exp_other_k}
For $k=1$, we prove that the Reptile algorithm's gradient is only different from the expected gradient by a constant factor. However, with other choices of $k$, the parameters may converge to a completely different point. In this section, we briefly present results for $k=2$ in the training of both meta models, and examine the how the results differ from the choice of $k=1$. The change of MAE and MSPR metrics, as well as the IWP metric are listed in Table~\ref{tab:diff_k} (we only present video 1 in experiment 1 for ease of presentation). We use the same partial adaptation configuration as in Section~\ref{sec:improve_vp}. Similar to Table~\ref{tab:perf_metrics}, the $\Delta$MAE and $\Delta$MSPR values denote the min-max change followed by the average change, compared with the single global model approach.

We find that the meta models trained with $k=1$ and $k=2$ perform similarly (details omitted due to the limit of space). Table~\ref{tab:diff_k} demonstrates that our algorithm is still effective for a choice of $k$ other than 1.

\begin{table*}
  \caption{List of Main Hyperparameters}
  \label{tab:hyperparams}
  \small
  \begin{tabular}{ccccccccc}
    \hline
     & $\alpha$ & $\eta^\text{(d)}$, $\eta^\text{(a)}$ & $\epsilon^\text{(d)}$, $\epsilon^\text{(a)}$ & $\mu^\text{(d)}$, $\mu^\text{(a)}$ & $N^\text{(d)}$, $N^\text{(a)}$ & $\mathcal{M}_{\theta}^\text{(d)}$, $\mathcal{M}_{\phi}^\text{(a)}$ \#Training Iter & Time Interval & Sequence Length\\
    \hline
    Simulation & $\pi/8$ & 0.1, 0.1 & 0.1, 0.1 & 0.001, 0.001 & 10, 10 & 200, 200 & 0.1~s & 100\\
    Emulation & $\pi/8$ & 0.1, 0.1 & 0.1, 0.1 & 0.001, 0.001 & 10, 10 & 200, 200 & 1~s & 20\\
    \hline
\end{tabular}
\end{table*}

\begin{table}
  \caption{Performance Metrics for $k=2$ (Video 1 in Experiment 1)}
  \label{tab:diff_k}
  \normalsize
  \begin{tabular}{cccc}
    \hline
     Adaptation & $\Delta$MAE (\degree{}) & $\Delta$MSPR (\%) & IWP (\%)\\
    \hline
     Partial & $[-20, 1]$, -3 & $[-1, 55]$, 10 & 55\\
     Full & $[-20, 0]$, -3 & $[-1, 55]$, 10 & 55\\
    \hline
\end{tabular}
\end{table}

\section{System Prototype}\label{sec:sys_prototype}

\subsection{System Description}
To better understand the performance of the proposed algorithm, we build an \textit{emulation} system on one Oculus Quest 2 device. 
Due to the difficulty and lack of support for building customized video streaming applications on Oculus Quest 2 devices, we adopt the following approach. 
%do not implement the video rendering and display functionalities on the devices.
We store a copy of the video locally on the device, and do the following in a simultaneous fashion: 1) we keep playing the local copy of the video, and 2) at the same time, we emulate the video streaming process by proceeding according to Algorithm~\ref{alg:robust_vp}. The streaming is emulated as if the video is received from the server and rendered to the display (the received data are actually discarded), and the input for prediction is retrieved through the playing of the video's local copy in a time-synchronized manner.

The emulations suffice to study most of the aspects of our streaming system, including the bandwidth consumption, and other metrics such as MAE and MSPR. Moreover, by keeping a (high-resolution) local copy of the video always playing, we avoid the situation where users watch a very low-resolution video for a period of time or even experience video stalls. When these happen, the users' behaviors can be undefined, which can negatively impact the accuracy of our evaluation.

\subsection{Setup}
In the emulations, we use one 360\degree{} videos from Vimeo\footnote{\url{https://vimeo.com/}} that is a scenic recording of a cascade and its surroundings for 120 seconds.

The video is projected onto a two-dimensional plane with equirectangular projection. We tile the video into a $16\times16$ grid and chunk it into 1-second clips with open-source tools\footnote{Available at \url{https://github.com/gpac/gpac/wiki/Tiled-Streaming}}. The tiling and chunking are much coarser than the idealized assumptions in Section~\ref{sec:simulations}. However, this is the most fine-grained we can perform on the given videos due to the inherent limitations of the video and the tools. The resulting files are in the format compatible with dash.js\footnote{\url{https://github.com/Dash-Industry-Forum/dash.js.}}, where each of the files corresponds to a time stamp and a position within the video. The position is mapped with equirectangular projection to find the correct tiles.

The meta model training and adaptation are similar to the schedule in Section~\ref{sec:simulations}. The hyperparameters used in the emulation section can be found in Table~\ref{tab:hyperparams}. For this video, we use all users' head-tracking data from the dataset~\cite{wu2017dataset}, accommodated to the hyperparameters chosen in the emulation, to train the meta models. We use a fixed prefetch angle---the closest $3\times3$ tiles around the viewing direction---in the emulations due to the coarse tiling of the video.

We evaluate the performance with two types of motions: 1) (Calm) staying relatively calm and focusing on a ROI, and 2) (Random) randomly browsing within the video.

\subsection{Results}

% \begin{figure}[t]
%     \centering
%     \includegraphics[width=0.95\linewidth]{figs/network_speed.png}
%     \caption{Network Capacity Over Time.}
%     \label{fig:network_cap}
% \end{figure}

% \paragraph{Environment measurement} We externally measure the network speed between the server and a user device (Figure~\ref{fig:network_cap}), and we use the measured speed as the network capacity over time.

\begin{table}
  \caption{Performance Metrics for the Emulation}
  \label{tab:emulation}
  \normalsize
  \begin{tabular}{cccc}
    \hline
     Pattern & Approach & MAE (\degree{}) & \#Missing Tiles \\ %& Data (MB)\\
    \hline
     \multirow{2}{*}{Calm}  & Single Model & 2.3 & 78 \\ %& 76.6\\
     \cline{2-4}
     & \textbf{Ours} & 1.2 & 78 \\ %& 76.6\\
     \hline
     \multirow{2}{*}{Random} & Single Model & 28.0 & 1125 \\ %& 76.6\\
     \cline{2-4}
     & \textbf{Ours} & 27.6 & 1074 \\ %& 76.5\\
    \hline
\end{tabular}
\end{table}

We compare our approach with the single global model approach in both head movement patterns in Table~\ref{tab:emulation}. The metrics are the MAE, and the number of missing tiles in streaming. The number of missing tiles indicates the amount of data received by the user, assuming all tiles have equal sizes\footnote{We do not directly use the size of transmitted data because we observe there is a large randomness in the video tile sizes.}. The performance gain in the emulations is relatively smaller than those presented in the simulations. For instance, in the case of calm pattern, the MAE is already small, and thus, decreasing the MAE by nearly a half does not decrease the number of missing tiles due to the coarse tiling of video. In the case of random pattern, we  decrease the number of tiles transmitted by around 4.5\%. Nevertheless, our algorithm is demonstrated to outperform the baseline case according to the experimental results.

% \begin{table*}
%   \caption{Performance Metrics of Two Experiments for Both Partial and Full Adaptation}
%   \label{tab:compare_cls}
%   \small
%   \begin{tabular}{cccccccccccc}
%     \hline
%      Experiment & Metric &Approach  & Video 1 & Video 2 & Video 3 & Video 4 & Video 5 & Video 6 & Video 7 & Video 8 & Video 9\\
%      \hline
%      \multirow{4}{*}{Experiment 1} & \multirow{2}{*}{MAE (\degree{})} & CLS & 28 & 41 & 33 & 18 & 23 & 35 & 41 & 34 & 44 \\
%      \cline{3-12}
%       & & \textbf{Ours} & 28 & 41 & 38 & 26 & 34 & 29 & 30 & 37 & 39\\
%       \cline{2-12}
%       & \multirow{2}{*}{MSPR (\%)} & CLS & 82 & 80 & 84 & 91 & 90 & 73 & 60 & 81 & 77\\
%       \cline{3-12}
%       &  & \textbf{Ours} & 91 & 95 & 96 & 91 & 93 & 97 & 95 & 95 & 92\\
%       \hline
%     \multirow{4}{*}{Experiment 2} & \multirow{2}{*}{MAE (\degree{})} & CLS & 9 & 10 & 23 & 19 & 12 & 43 & 14 & 16 & 11 \\
%      \cline{3-12}
%       & & \textbf{Ours} & 11 & 20 & 18 & 24 & 11 & 33 & 16 & 19 & 17 \\
%       \cline{2-12}
%       & \multirow{2}{*}{MSPR (\%)} & CLS & 96 & 95 & 86 & 95 & 92 & 61 & 91 & 90 & 96\\
%       \cline{3-12}
%       &  & \textbf{Ours} & 94 & 91 & 96 & 93 & 94 & 96 & 92 & 92 & 94\\
%       \hline
% \end{tabular}
% \end{table*}

\section{FLOPs Computation}\label{appendix:flops}
We take the simplest ``neural network'', a dot product, as an example:
$
\mathbf{y}=\mathbf{x}\cdot \mathbf{w}\,\,,
$
where $\mathbf{w}\in\mathbb{R}^{n_\text{in}\times n_\text{out}}$ is the weight, $\mathbf{x} \in \mathbb{R}^{N\times n_\text{in}}$ is the input and $\mathbf{y}\in\mathbb{R}^{N\times n_\text{out}}$ is the output.

In the forward pass, the FLOPs is $2 N n_\text{in} n_\text{out}$. In the backward pass, the FLOPs for the gradient computation of the weight and the input are both $2 N n_\text{in} n_\text{out}$. Therefore, the total FLOPs for the backward pass is $4 N n_\text{in} n_\text{out}$, doubling that in the forward pass. The FLOPs for both forward and backward passes is $6 N n_\text{in} n_\text{out}$. Roughly speaking, the total FLOPs of both passes triples the FLOPs in the forward pass. This discussion is a guideline for the computation of FLOPs in models involving more complex operations.

\end{document}